\documentclass[11pt,a4paper]{article}
\usepackage[T1]{fontenc}
\usepackage[utf8]{inputenc}
\usepackage{mathptmx}
\usepackage[margin=24mm]{geometry}
\usepackage{amsmath,amssymb}
\usepackage{graphicx}
\usepackage{array,tabularx,booktabs}
\usepackage[numbers,sort&compress]{natbib}
\usepackage{placeins}
\usepackage[font=small,labelfont=bf]{caption}
\usepackage[hidelinks,breaklinks]{hyperref}
\hypersetup{pdftitle={Human-Supervised Artificial Intelligence Researchers in Electric Power Systems: Knowledge, Physics, Simulation, and Governance for Large Language Models},pdfauthor={Xiangrui Meng, Jingcheng Liu, Ke Wang, Miaopeng Yu, Zhengping Lin}}
\title{Human-Supervised Artificial Intelligence Researchers in Electric Power Systems: Knowledge, Physics, Simulation, and Governance for Large Language Models}
\author{\normalsize Xiangrui Meng \quad Jingcheng Liu \quad Ke Wang\\[3pt]
\normalsize Miaopeng Yu \quad Zhengping Lin\\[7pt]
\small China Southern Power Grid Research Institute Co., Ltd.\\
\small Guangzhou 510663, China}
\date{}
\begin{document}
\maketitle
\begin{abstract}
Conclusions in power-system research require a clear evidential basis. Researchers need to understand the data sources, modelling assumptions, and solution procedures to assess the conditions under which a conclusion holds. Large language models (LLMs) can help researchers extract information from documents and translate research requests into code and tool calls, but their outputs still require physical validation. Focusing on human-supervised artificial intelligence (AI) researchers, this review examines power-system LLM applications and related scientific workflows, together with physics-grounded power AI studies, to discuss the conditions for using LLMs in scientific research. In the proposed knowledge--physics--simulation--governance framework, LLMs translate information and coordinate tasks, simulators and validation tools test models, and experts retain authority over critical assumptions, safety boundaries, and final conclusions. Research processes should preserve materials for review so that claims can be traced to their sources, computations can be reproduced, and failed attempts and approvals can be examined. On this basis, a staged development roadmap and requirements for research reporting are proposed. Evaluation should focus on research artifacts and their validation to assess whether LLMs can improve the efficiency of power-system research within clearly defined permissions.
\end{abstract}

\section{Introduction}
The large-scale integration of renewable generation and energy storage, together with changes in electricity demand and digital control, has made power-system modelling, computation, and result validation more difficult. At the same time, the data, standards, and models used in a study are often scattered across different documents and software systems. Checking a conclusion requires connecting these sources of evidence. Measurements must be consistent with simulation models, while modelling assumptions often come from standards, operating rules, or expert judgement. Researchers therefore need to establish whether the equations, parameters, and software environment are appropriate for the research question.

Artificial intelligence (AI) is widely used for specific power-system tasks. Deep neural networks can forecast electricity demand and renewable generation, estimate operating states from measurements, or identify faults. Graph neural networks (GNNs) emphasize the representation of electrical connections between devices. Reinforcement learning explores control strategies, such as deciding when a battery should charge or discharge within operating limits. Physics-informed learning further incorporates equations and constraints into training. \citep{ref016,ref017,ref019,ref021,ref022,ref156,ref157,ref179,ref041,ref042,ref043} These methods generally construct predictors, surrogate models, or controllers for predefined datasets and objectives, with evaluation focused on model performance. A complete research process must also explain how assumptions were formed, how constraints entered the model, and how computational results support their interpretation. Reviews of probabilistic learning and smart-grid AI provide broader accounts of existing tasks and methods. \citep{ref132,ref141,ref070,ref072} Table~\ref{tab:literature} summarizes the three literature streams used in this review and their roles.

The role of large language models (LLMs) is gradually extending to the organization of scientific work. Researchers can express requirements in natural language, allowing a model to consult documents, generate code, and call simulation or optimization tools. Existing studies have examined this approach in energy systems, including load forecasting and distribution-network dispatch, as well as markets, cybersecurity, and reliability. Digital twins, power electronics, and power-system question answering have also been explored. \citep{ref001,ref002,ref003,ref004,ref007,ref009,ref012,ref013,ref014,ref015,ref120,ref121,ref143} In these studies, LLMs begin to help clarify questions, organize retrieved evidence, and use tool feedback to revise code or interpret results, potentially connecting research steps that were previously relatively separate.

However, introducing LLMs into research also allows errors to propagate through workflows. A fluent answer may still violate circuit laws, overlook equipment limits, or omit post-fault security requirements. Even executable code may solve an optimization problem different from the one intended. Retrieved standards may be inapplicable because they concern a different version, region, or voltage level. If an agent system checks only whether the final task was completed, unreasonable assumptions in intermediate steps can remain hidden. The process through which a language model handles evidence must therefore be examined to determine whether that evidence supports the final power-system conclusion.

In this review, an AI researcher is a system that assists with research tasks within explicit permissions and preserves the evidence underlying the work. Starting from a research request, it can retrieve evidence, formulate candidate assumptions, translate models into scripts, and call simulation tools. After obtaining results, it can help analyze failures, compare alternatives, and prepare a report for expert review. Decisions about modelling choices, safety boundaries, and final conclusions remain with experts. On this basis, this review examines power-system LLM research, considers which workflow designs from artificial intelligence for science (AI-for-science) can be adapted, and draws on physics-grounded power AI to establish the validation requirements those designs must satisfy. Using verifiable research artifacts as the basis for evaluation, it proposes a knowledge-physics-simulation-governance framework and examines the responsibilities and limitations of LLMs at each stage. A staged roadmap and reporting requirements further identify the evidence and controls needed as the scope of agent tasks expands.

\begin{table}[tbp]
\centering\small
\caption{Research focus and roles of the three literature streams.}
\label{tab:literature}
\setlength{\tabcolsep}{4pt}\renewcommand{\arraystretch}{1.18}
\begin{tabularx}{\textwidth}{>{\hsize=0.66\hsize\linewidth=\hsize\raggedright\arraybackslash}X>{\hsize=1.5\hsize\linewidth=\hsize\raggedright\arraybackslash}X>{\hsize=0.84\hsize\linewidth=\hsize\raggedright\arraybackslash}X}
\toprule
\textbf{Literature stream} & \textbf{Main research content} & \textbf{Role in this framework} \\
\midrule
Power-system LLM studies & Review LLM and generative-AI applications in energy and grids \citep{ref001,ref002,ref003,ref118,ref129,ref130,ref125,ref126,ref144,ref148}; study modelling, simulation, optimization, digital twins, regulatory compliance, markets, and design \citep{ref012,ref013,ref014,ref015,ref120,ref121,ref113,ref119,ref124,ref133,ref140,ref151}; propose power-agent architectures and controller-design concepts \citep{ref143,ref150} & Provide direct application evidence to examine which research tasks language models can support and how those tasks connect \\
AI-for-science workflows & Discuss research agents and scientific discovery \citep{ref029,ref030,ref031,ref051,ref108,ref216}; explore reasoning, hypothesis generation, and preliminary experiments through retrieval, knowledge representation, and tool execution \citep{ref028,ref026,ref048,ref052,ref093,ref103,ref197,ref218,ref219,tang2025chemagent,tang2026eigenagent} & Provide workflow design references for organizing multistep tasks, handling execution feedback, and arranging human review \\
Physics-grounded power AI & Study state estimation, optimal power flow, and power-model validation through physics-informed learning and topology modelling \citep{ref016,ref017,ref019,ref021,ref022,ref156} & Establish validation requirements for equations, network connectivity, constraints, and numerical solution procedures \\
\bottomrule
\end{tabularx}
\end{table}

\section{Responsibilities and boundaries of power-system AI researchers}
AI-for-science research has begun to involve LLM agents in planning and executing scientific workflows. Related systems have attempted to connect hypothesis generation, experiment or code execution, and result analysis, while recording the process and drafting reports. Tool-augmented reasoning and knowledge graphs have also been used to support these tasks. \citep{ref030,ref031,ref216,ref026,ref052,ref093,ref197,ref218,ref024} For power systems, further questions concern how these systems comply with grid physics and who takes responsibility for engineering judgement. PowerAgent has proposed a conceptual architecture and roadmap for the power domain, \citep{ref143} but their effectiveness still requires validation in specific research tasks. Power-system research may have operational consequences and is subject to industry regulation, so general scientific demonstrations do not establish how much authority an agent should have. Broader reviews discuss scientific uses of foundation models and the design of planning, memory, and tool interfaces, providing references for further research. \citep{ref108,ref191,ref194,ref046,ref032}

When applying a general AI research system to power systems, its permissions must first be defined. A complete-looking report does not establish that its conclusions are reliable; the underlying assumptions and models still require domain-specific checks. Circuit equations and equipment limits cannot be relaxed arbitrarily because a problem is difficult to solve, and dynamic models must be appropriate for the disturbance under study. If a workflow uses an invalid operating point, ignores solver warnings, or presents an exploratory run as a conclusion, the study may be invalid even if every step was executed.

The distinction between power flow and dynamic stability analysis illustrates this risk. Power-flow calculations determine steady-state bus voltages and line power flows for given generation, demand, and network connectivity. Dynamic stability analysis examines whether the system can maintain or recover an acceptable state after a disturbance. Power-flow results alone cannot establish post-disturbance dynamic stability. Similar errors include a case whose network connectivity differs from the assumptions, or mixed units and normalization bases. Optimal power flow (OPF) seeks an operating plan that minimizes an objective such as cost subject to network and equipment constraints. Omitting a constraint directly changes the feasible set. Once such errors enter the code, they can propagate through simulation results into reports and engineering judgement.

An AI researcher should act within the scope authorized by researchers. It can organize evidence, prepare model drafts, generate scripts, run solvers, and submit results and failure analyses for review. Researchers define the problem, approve modelling assumptions, and authorize data use. Agents should not independently set safety boundaries, release research claims, or make grid-operating decisions.

These responsibilities must be implemented through system design. Information processing should preserve sources so that parameters and assumptions can be traced to the original materials. Generated models should undergo the corresponding physical checks and simulation tests. During code execution, a sandbox should limit access to data and tools, and steps involving critical judgements should pause for human approval. Responsibility remains with researchers and institutions. Expanding agent autonomy is justified only when the evidence can be reviewed and tool behavior can be controlled.

System maturity also depends on recognizing when execution should not continue. Missing data or unsupported assumptions require additional evidence. An unsuitable tool or an infeasible case requires an explanation and reconsideration of the task. Requests that exceed permissions or involve safety judgements should be referred to a human and, where necessary, refused. These responses help researchers determine which results can withstand scrutiny.

\begin{figure}[tbp]
\centering
\includegraphics[width=\textwidth]{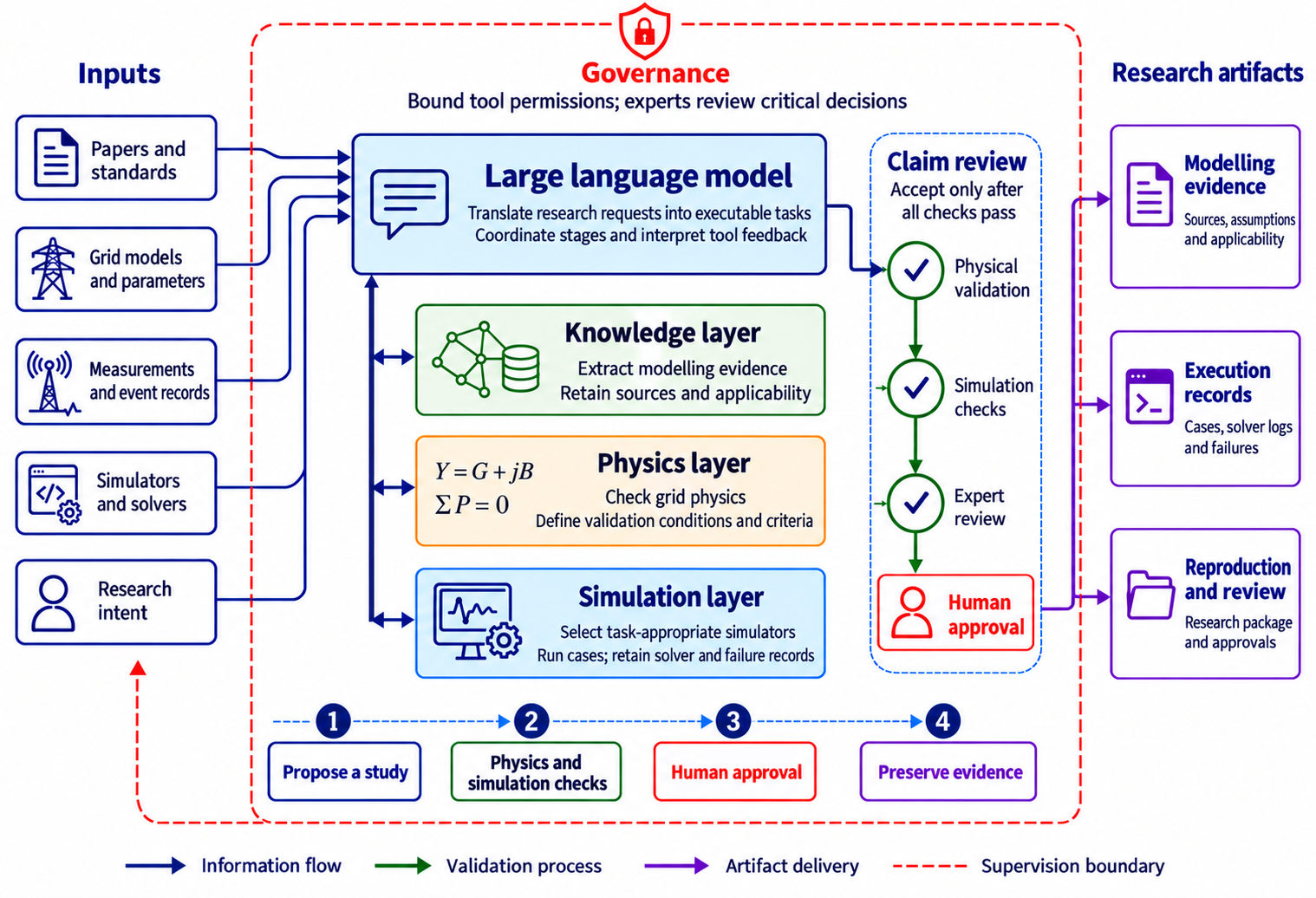}
\caption{Knowledge-physics-simulation-governance framework for human-supervised AI researchers in electric power systems.}
\label{fig:framework}
\end{figure}

Figure~\ref{fig:framework} presents the corresponding framework. The LLM translates researchers' requests into executable tasks and coordinates information exchange between the knowledge, physics, and simulation layers. The knowledge layer supplies modelling evidence, physical models and simulators test candidate solutions, and researchers review critical decisions. Governance spans these activities, limits operational permissions, and preserves the basis for review. The materials retained at each stage together provide the evidence underlying the study, enabling experts to locate the origins of errors without relying solely on the final report.

\section{Related research and its implications}
Starting from the practical needs of power-system research, this review selects power-system LLM studies, AI-for-science workflows, and physics-grounded power AI as its basis for discussion. Table~\ref{tab:literature} summarizes the content and roles of these three streams. The analysis focuses on how LLMs participate in modelling, tool execution, and result interpretation, and on the conditions under which the methods apply to power systems. Because studies use different tasks, datasets, and test cases, this review compares methods and validation approaches without constructing a unified ranking of model performance or engineering deployment maturity.

The three streams contribute application experience, workflow design methods, and validation requirements to the framework. Power-system LLM studies show how language models enter specific tasks. AI-for-science research offers approaches to organizing multistep work, such as using retrieval and tool feedback to advance a study and recover from execution failures. \citep{tang2025agentkb} Establishing whether a generated model follows grid physics requires physics-grounded power AI to identify checks on equations, topology, and numerical solution procedures. Traditional power-AI reviews provide the background of established methods and evaluation. Organizing the literature by these uses helps distinguish the transferability of cross-domain methods from their applicability to power systems.

Existing power-system LLM research covers forecasting, energy management, cybersecurity, markets, digital twins, and question answering. \citep{ref009,ref012,ref015,ref124,ref010,ref112,ref114,ref122,ref123,ref154} In modelling and tool execution, studies have attempted to generate dispatch models or grid cases from natural-language requirements, improve simulation through execution feedback, assess dispatch-regulation compliance through retrieval-augmented generation (RAG), and organize market-agent tasks. \citep{ref013,ref014,ref120,ref121,ref133,ref151} In specific applications, building-control studies use LLMs to explain feature-attribution results, \citep{ref011} reinforcement-learning-assisted optimization \citep{ref113,ref119} and distribution-network reconfiguration \citep{ref145} focus on generating and improving operating plans, and power-electronics research explores retrieval-assisted design and proposes controller-design agents. \citep{ref144,ref140,ref150} Perspectives on grid operation and resilience also discuss directions for further integration. \citep{ref036,ref115,ref128} These studies inform different aspects of LLM participation in power-system research. Explanatory accuracy, operational feasibility, and the reliability of a complete research process still require separate evaluation.

AI-for-science research offers specific workflow mechanisms for adaptation. Related systems attempt to extract information from literature, use knowledge representations to support hypothesis generation, call tools or execute experiments, and draft reports from the results. \citep{ref028,ref026,ref052,ref197,ref218,ref024,tang2026cellforge} Reviews and perspectives further discuss how to organize these operations in scientific and engineering research. \citep{ref030,ref031,ref216,ref055} Applying these methods in power systems requires additional domain-specific validation and explicit limits on execution permissions. Biomedical and chemistry perspectives provide further design references. \citep{ref029,ref051} Physics-guided foundation models in water science offer a non-LLM example of combining learned and physical models. \citep{ref034}

Physics-grounded power AI informs validation design even though many of these studies do not use LLMs. State estimation requires correct treatment of measurements and network topology. OPF graph neural networks must represent different node or edge types, while dynamic surrogate models must be tested against disturbance responses. Transient stability assessment and reinforcement learning for grid operation also have their own conditions of applicability. \citep{ref016,ref017,ref019,ref021,ref156,ref179,ref018,ref020,ref158,ref160,ref165,ref168,ref170,ref171,ref172,ref173,ref174} Together, these studies show that power-system models require task-appropriate tests, including checks on equations and solution procedures and reporting of uncertainty. Stating physical requirements in a prompt cannot replace numerical checks.

\begin{figure}[tbp]
\centering
\includegraphics[width=0.78\textwidth,height=0.68\textheight,keepaspectratio]{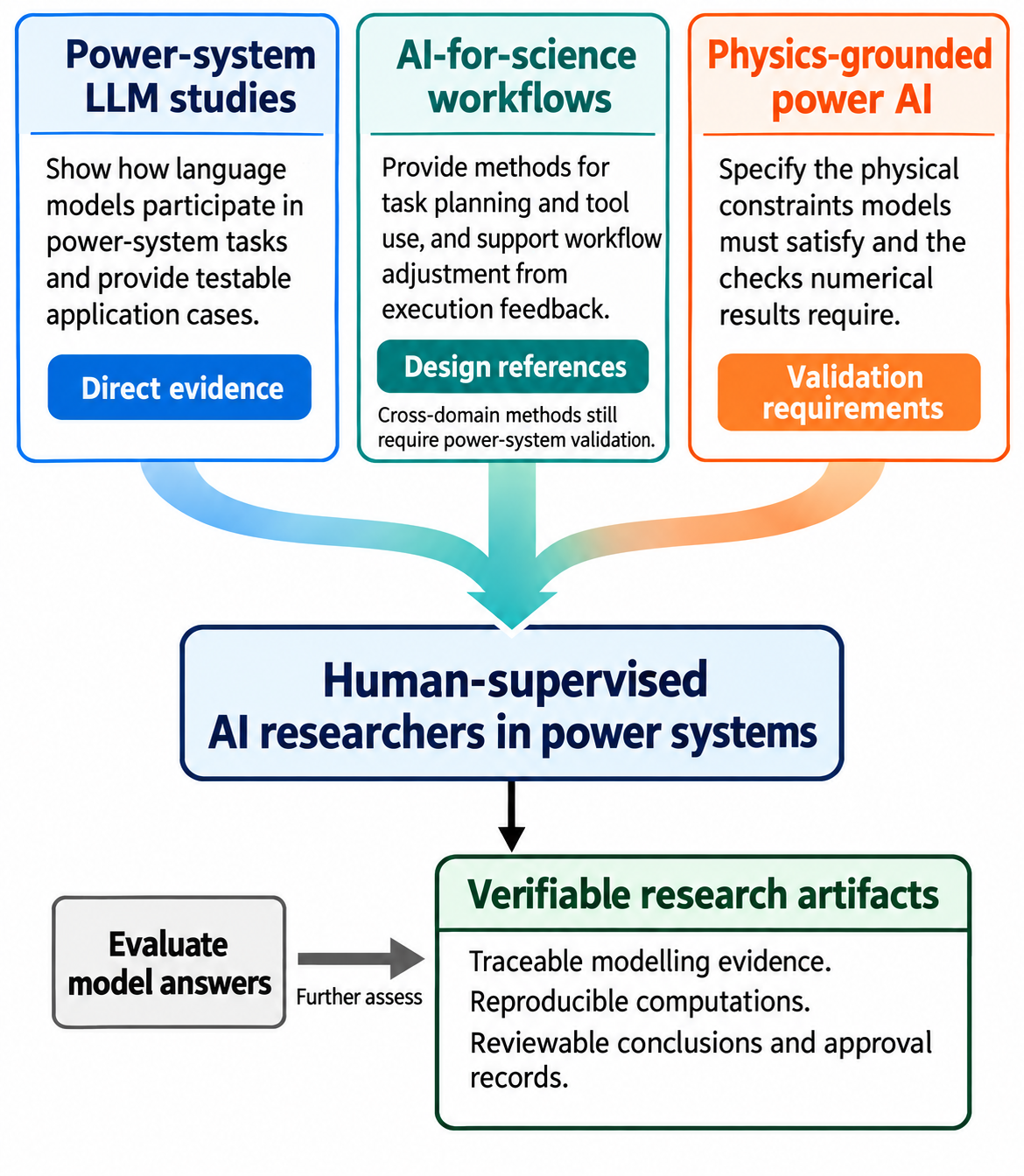}
\caption{Contributions of the three literature streams to the research framework. Power-system LLM applications provide direct evidence, cross-domain scientific workflows offer design references, and physics-related research establishes validation requirements. Together, they support the evaluation of research artifacts.}
\label{fig:evidence}
\end{figure}

Table~\ref{tab:literature} sets out the research content of the three literature streams, and Figure~\ref{fig:evidence} shows how they jointly support the research framework. Evaluation must further examine the materials retained by a workflow and establish whether they support source tracing, computational reproduction, and review of conclusions.

\section{Research artifacts and their evaluation}
Trustworthy research artifacts are materials and records retained during a study that other researchers can inspect. They should allow others to understand assumptions, rerun computations, and assess whether conclusions hold. Conversations, prompts, and performance tables provide some information, but they must be linked to sources and executable cases and must identify which results have been reviewed. Evaluation of power-system LLMs should therefore examine what the workflow leaves behind and whether those materials help expose errors or refute an original claim.

Research artifacts can be grouped by purpose into evidence artifacts, modelling artifacts, execution records, and governance records. Evidence artifacts retain the original text and context supporting a claim, such as a standard clause or an event report. Modelling artifacts explain how that evidence enters the model, define variables and units, and describe network connectivity, operating assumptions, and uncertainty. Execution records preserve the cases and scripts actually used, together with solver settings and outputs, including convergence logs, residuals, and failed runs. Governance records identify who authorized or approved each step and retain data-access decisions, permission settings, prompt-injection checks, and human interventions.

Review requires connecting these materials. Even a claim supported by the literature must be translated into explicit modelling assumptions before it can be tested numerically. Reproducing an OPF result requires both the formulation and solver settings. Solver logs should be linked to input data to establish whether the computation addressed the original research question, while approval records identify where experts exercised judgement. LLM-assisted computer-aided design (CAD) and design customization demonstrate the use of structured outputs, \citep{ref097,ref211} and research on prompt patterns and non-expert prompting also supports recording how user requirements become outputs. \citep{ref050,ref202} By connecting these materials, researchers can distinguish errors caused by misreading evidence, modelling omissions, execution failures, or governance failures. Table~\ref{tab:artifacts} lists the corresponding checks.

Research artifacts should also facilitate challenge. Reviewers should be able to locate a claim's source, identify assumptions that could change a result, and select methods that can test the claim. Excluded failed runs and the person who approved the final interpretation should also be recorded explicitly. These details provide a basis for independent review.

Developing from an LLM assistant into an AI researcher means gradually taking on more verifiable research operations. Early assistants primarily explain concepts or draft code. Building on this, a system can extract source-linked evidence and represent assumptions as typed data structures. Connection to simulation tools then enables it to generate cases, read errors, and propose repairs. Planning experiments and comparing multiple runs further requires a complete execution history and negative results. A human-supervised AI researcher coordinates these processes under expert authorization. Its maturity should be assessed by whether the research process can be reconstructed and its conclusions can withstand testing.

The following sections discuss how these materials are produced. The knowledge layer extracts assumptions from original sources, the physics layer translates them into testable models, and the simulation layer executes those models to obtain results. Governance rules span the process and specify permissions and points of human intervention.

\begin{table}[tbp]
\centering\small
\caption{Categories of research artifacts in LLM-enabled power-system research.}
\label{tab:artifacts}
\setlength{\tabcolsep}{4pt}\renewcommand{\arraystretch}{1.18}
\begin{tabularx}{\textwidth}{>{\hsize=0.6\hsize\linewidth=\hsize\raggedright\arraybackslash}X>{\hsize=1.2\hsize\linewidth=\hsize\raggedright\arraybackslash}X>{\hsize=1.2\hsize\linewidth=\hsize\raggedright\arraybackslash}X>{\hsize=1.0\hsize\linewidth=\hsize\raggedright\arraybackslash}X}
\toprule
\textbf{Artifact category} & \textbf{Power-system examples} & \textbf{Required validation} & \textbf{Consequences if absent} \\
\midrule
Evidence artifacts & Retain original standard clauses or event reports; link data sources and literature claims to their citation context & Are sources recoverable and versions and access conditions clear? Does their scope support the claim? & Unsupported statements enter model design \\
Modelling artifacts & Specify network connectivity and demand assumptions, units and normalization bases, study conditions, and applicable rules & Are variable types and units consistent? Have boundary conditions, applicability, and uncertainty been confirmed? & Undocumented assumptions change the actual research question \\
Execution records & Preserve actual cases, scripts, solver settings, and outputs; archive residuals, convergence logs, and failed runs & Are data and dependency versions available? Are solver tolerances, residuals, and constraint violations consistent with execution status? & Computations are hard to rerun or debug and results cannot be tested adequately \\
Governance records & Record data authorization, operational permissions, tool calls, and human interventions; explain rejected assumptions and claim approvals & Are approvers identified? Are credentials and sensitive data controlled, and have prompt-injection checks been performed? & Responsibility within the workflow is unclear \\
\bottomrule
\end{tabularx}
\end{table}

\section{Evidence extraction and assumption organization in the knowledge layer}
Many modelling choices originate in documents. Equipment manuals specify ratings and operating limits; grid codes and operating procedures establish applicable conditions; and protection-setting documents and market rules influence the definition of specific research questions. Incident reports, project documents, and existing papers or patents may also support assumptions. If these materials appear only as background text, readers cannot determine how model parameters and boundaries were established. LLMs can help extract requirements and claims, link them to structured modelling artifacts, and retain their authoritative sources.

Materials-chemistry research has attempted to extract entities and relations from scientific text, \citep{ref024} while preliminary workflows for scientific facilities combine document retrieval with tool use. \citep{ref028} These methods can inform the design of a power-system knowledge layer, but their domain applicability still requires validation. For example, after identifying a generator or line in equipment documents, a system can record its connections and operating limits. Events extracted from fault reports can be linked to the corresponding responses. The resulting candidate assumptions must specify parameter meanings, model boundaries, and uncertainty, with information types and sources recorded. Domain-specific language representations such as MatSciBERT help identify technical terms, \citep{ref066} while knowledge graphs and corpus platforms can organize extraction results. \citep{ref023,ref209,ref212} Structuring information does not remove the need to assess whether the assumptions are valid in engineering terms.

Preserving a source also requires retaining the conditions that determine its applicability. A standard clause must identify its version and applicable region, and a requirement concerning a voltage level must remain within its original scope. Parameters should be recorded with units, base values, and measurement conditions. A paper's claim should be identified as arising from an experiment, simulation, review, or speculation. If documents conflict because they concern different equipment classes or periods, the respective conditions should be retained. Forcing them into one definite answer hides ambiguity in the original evidence.

The authority of a source also determines which judgements it can support. A grid code or operating procedure may establish a mandatory boundary, a manufacturer's manual may specify equipment-use conditions, and a peer-reviewed simulation result holds only under its associated assumptions. An event report records observations that occurred, whereas informal design notes may only propose ideas for testing. Feeding these materials to a model as equivalent text fragments can confuse normative requirements with research assumptions. A RAG workflow should retain these distinctions and explain which modelling operations each source can support.

Retrieved information in power-system RAG should be able to affect subsequent modelling. After retrieving a requirement, the system must check its applicability to the current problem and use it to define variable meanings, parameter ranges, or constraints, adding validation checks where necessary. Retrieval can then both substantiate an existing claim and expose judgements that still require testing. Question answering using domain-specific application programming interfaces (APIs), manufacturing retrieval, enterprise benchmarks, and energy chatbots demonstrate ways to supply task context, \citep{ref027,ref090,ref109,ref138} but whether the information can become an engineering constraint depends on verification of its scope.

Dynamic knowledge graphs have been used to coordinate distributed laboratory workflows and record data provenance, \citep{ref053} providing a design precedent for evidence organization in power-system research. A knowledge graph can represent relations among equipment, events, and constraints and link them to clauses, datasets, or experiments. Retrieval indexes retain original passages for review, while structured databases store cases, tool calls, and solver settings. LLMs can help query and translate information between these systems. Each transformation should record its sources and any uncertainty it may introduce. The corresponding storage systems retain the evidence so it can be examined independently of the language model's answers.

Knowledge-graph RAG may fail in contextual understanding, relation mapping, or disambiguation. \citep{ref039} In power systems, a retrieved document may be outdated or apply to a different region or voltage level. Unit-conversion errors and misread topology relations can also affect model construction. If a provisional inference is passed to a simulator as fact, even numerically normal results may rest on a false premise. Literature-synthesis tools, battery-recycling document agents, and PowerGridQA provide settings for evaluating document processing and question answering, \citep{ref154,ref106,ref139} but these evaluations cannot replace checks on executable grid models. Further work must establish whether retrieved information remains valid after it becomes assumptions and constraints.

This raises reliability questions along the path from retrieval to modelling: whether different names for the same device can be matched, whether a clause's version and scope can be determined, and whether textual conditions and exceptions can be translated fully into model constraints. Checking only whether an answer cites the correct document is unlikely to expose omissions in the transformation. The original text, extracted assumptions, and actual case must be examined together. When sources conflict or conditions are incomplete, the system should preserve ambiguity and request additional information instead of directly generating a unique parameter value. Once an error reaches simulation, distinguishing retrieval errors, assumption-transformation errors, and solver-setting errors also affects repair. Parameter changes must return to their supporting evidence, and changes to research boundaries require renewed review. Evaluation should therefore cover these transformations and test whether a system can locate errors and stop when evidence is insufficient.

\section{Modelling and validation in the physics layer}
A power-system model must satisfy the physical requirements of the question under study. Currents and voltages obey circuit laws, transmitted power is limited by equipment ratings, and disturbance responses depend on dynamic models. An LLM can assist in proposing a candidate model, but accepting it requires the corresponding physical and numerical tests.

Physics-grounded power AI shows that different models require different checks. Physics-informed learning incorporates differential equations, conservation laws, or boundary conditions into training, while topology-aware graph models emphasize electrical connectivity. OPF surrogates require tests of the feasibility of returned solutions. Dynamic surrogates must be tested at different time scales and under different disturbances. State estimation must establish whether measurements are sufficient, explain dependence on topology, and report residuals and uncertainty. \citep{ref016,ref017,ref019,ref156,ref020,ref168,ref170,ref174,ref164} Reviews of inertia estimation, residual dynamic-parameter models, and Bayesian physics-informed identification further show how model choice and parameter uncertainty affect interpretation. \citep{ref069,ref075,ref086,ref082} Mathematical models generated by LLMs should undergo these checks as well. Physical prompts in the generation process do not establish model validity.

During modelling, an LLM can translate a natural-language problem into variables and constraints, help draft a power-flow or OPF formulation, generate code, and call candidate datasets and solvers. Whether its output is usable depends on subsequent testing. A power-flow solution requires numerical checks, and a stability conclusion requires an appropriate dynamic model. Solver warnings and safety-related interpretations must also be reviewed. Validation must be able to reject generated results; otherwise errors in the model cannot be corrected effectively.

Validation should proceed alongside model execution. Before a run, inputs should be checked for consistency, particularly units and normalization bases, network connectivity, and parameter ranges, with timestamps and data sources confirmed. During execution, code and tool calls should be checked against the intended equations and constraints. After execution, task-specific checks should cover convergence, residuals, and limit violations. Dynamic or protection studies also require analysis of stability margins and protection actions and the handling of solver warnings. Counterfactual checks further change operating conditions, for example by introducing faults, extreme weather, or missing data, to examine whether conclusions remain supported. Market studies can similarly vary market conditions. Learned dynamic components, including PINNSim, plug-and-play simulation, and physics-informed network variants, also require their coupling assumptions and simulation conditions to be specified. \citep{ref081,ref076,ref077,ref085}

Choosing a validator is itself part of modelling judgement. For example, deterministic OPF alone cannot establish resilience to rare events, and interpreting a cyber event requires an explicit threat model. Learning-based OPF and solver warm-start studies address numerical tasks different from those of graph reconfiguration or load-margin prediction, so their checks are not interchangeable. \citep{ref166,ref163,ref084,ref080,ref083} Researchers should first identify the operational or economic question behind a claim, determine whether it involves protection actions or interactions between information and physical systems, and then choose suitable validation methods. This prevents validation from drifting away from the original research question.

Each research artifact should state the grounds for accepting or rejecting it. Power-flow studies must check circuit equations, voltage and equipment limits, and report solution accuracy. In addition to checking the returned solution, OPF studies must verify that the objective and constraints are complete. Dynamic studies should describe the disturbance and control assumptions and identify the criteria used to evaluate the response. Checks should be selected for the task, with reasons given when a check does not apply. Solver success establishes only part of what is required; it cannot replace an assessment of whether the model matches the research question.

Validation and interpretation in protection, market, and cybersecurity research also depend on their respective task conditions. Protection studies must identify the devices that act after a fault and the timing of those actions. Market analysis must specify trading and settlement rules. Cybersecurity studies must define what an attacker can access or alter and distinguish attacks on the language-model interface from attacks on grid measurements and control systems. In interpreting results, stability judgements require evidence such as response curves or protection actions. Forecasting analysis should distinguish the effects of weather and calendar factors from those of the model itself, while security analysis should distinguish observed signs from inferences about attacker intent. Studies of state estimation and its uncertainty \citep{ref087,ref161,ref169}, a review of frequency control \citep{ref180}, and microgrid-control research under communication failures \citep{ref190} all indicate that measurement conditions, time scales, and communication assumptions affect interpretation. Uncertainty must therefore be explained in relation to applicable conditions, with observations, inferences, and judgements distinguished.

\section{Execution and experimental feedback in the simulation layer}
Simulation tools should be selected for the research question. Steady-state power flow and OPF can use the corresponding network-computation tools, whereas distribution-network analysis or dynamic-response studies require simulators suited to their model characteristics. Optimization tasks also need modelling tools and solvers that execute the objectives and constraints correctly. A workflow should record why a tool was selected, describe the model it actually solves, and explain how outputs are checked. A software name identifies the execution environment but does not establish that the original research question has been tested.

After code generation, execution feedback gives a system the opportunity to detect and correct errors. A tool run provides solution status and result files and may also return warnings, residuals, or infeasibility diagnostics. Software-engineering reviews and scientific-programming studies offer methods for checking generated code, \citep{ref063,ref199,ref091,wang2025openhands,tang2024mlbench} but executable code must also be checked against the intended equations and numerical experiments. With simulation tools connected, an agent can run a case, use feedback to identify inconsistencies, propose changes, and rerun the computation. Assumption changes should follow the established approval rules, and repairs should be recorded fully so that subsequent experiments can explain why results changed.

Transformations from information sources to simulation results should be reconstructable. After information extraction from a research request and its source documents, typed assumptions should be formed and then translated into model constraints and tool inputs. When simulation or optimization has run, solver logs and failure records should be retained with the results for human review. Abandoned cases and negative results must be preserved alongside successful runs because they may reveal conditions under which assumptions fail or tools reach their limits of applicability.

Existing power-system studies have demonstrated several parts of this process. Automated dispatch for active distribution networks extracts information from natural-language requests, formulates a problem, and generates code. \citep{ref013} Tool-augmented grid-model generation allows users to construct cases through natural language. \citep{ref120} Feedback-driven simulation agents address DALINE and MATPOWER tasks by using retrieval and reasoning to organize tool execution and adjusting operations in response to errors. \citep{ref014} RAG-based assessment of dispatch regulations emphasizes compliance checking and human review, \citep{ref121} while market-agent research uses modelling and simulation to generate bidding strategies. \citep{ref133,ref151} Organizing these components into a scientific workflow still requires explicit accounts of how intermediate assumptions were approved and which checks each step passed. Task-completion records can support research judgement only when linked to this evidence.

Failed runs help identify the limits of a workflow. For example, power-flow non-convergence means that a numerical procedure failed to find a solution within its stopping criteria; it does not directly imply physical impossibility. OPF infeasibility concerns whether the stated constraints can be satisfied simultaneously, and solver diagnostics must be interpreted in the specific circumstances. Inconsistent topology, missing data, or unit errors can cause failures. Software-version differences and omitted constraints can also change results. Preserving these records allows researchers to determine whether a problem originates in assumptions, tools, or execution and where human intervention is required. Reporting only successful runs overestimates agent reliability and obscures a model's valid scope for subsequent research.

Failure classification should guide repair decisions. Syntax errors can usually undergo an initial attempt at automatic repair. Invalid cases require checks on inputs and their assumptions, whereas infeasible models and numerical non-convergence need separate diagnosis. Outputs that violate physical laws require a return to the model and its constraints. Unsupported assumptions require additional evidence, while unsafe tool requests should be refused or referred to a human. Different categories permit different changes. Root-cause analysis in cloud systems and industrial troubleshooting provide references for diagnostic design, \citep{ref105,ref196} and variable extraction from scientific-software documentation can support metamorphic testing. \citep{ref210} Recording error types and responses helps compare systematically the conditions under which an agent can recover.

A reproducibility package should allow other researchers to reconstruct how a study was executed and see what it actually validated. Recursive reasoning and refinement, constitutive-model discovery, and the experimentally tested Virtual Lab provide different levels of evidence. \citep{ref048,ref103,ref219} For power-system simulation, the package should preserve input data and cases, specify parameters, unit conventions, and tool versions, link prompts or task templates to the actual commands or API calls, and record random seeds. Result files should be archived with warnings, failed attempts, and human interventions. Reviewers can then determine which models and solver settings a conclusion depends on, and where prompt changes or human corrections affected the results.

\section{Collaboration between multimodal evidence and specialist models}
Power-system research combines evidence in different forms, including text, measurements, network structures, and images. Demand and generation measurements must be linked to equipment, and event logs must be interpreted with the network structure. Images and simulation outputs may also address the same question. The first step is therefore to establish that these materials refer to the same equipment, time interval, and operating conditions, and to check their assumptions. \citep{ref005} Once aligned, different data types still require suitable specialist models.

Time-series foundation models and forecasting methods with LLM-like architectures have been applied to load, wind power, building energy use, and multienergy tasks. \citep{ref007,ref112,ref114,ref122,ref123,ref008,ref127} They are candidate methods when samples are limited, local historical data are unavailable, or transfer across settings is needed. Forecasts remain affected by weather, energy-use behavior, grid connectivity, control policies, and market incentives. Good transfer performance does not establish that a model has captured these mechanisms, so forecast provenance and uncertainty must still be specified when forecasts enter simulation or decision analysis. Reviews of renewable-energy, photovoltaic, and Transformer forecasting provide background on baseline methods, \citep{ref188,ref189,ref177} including clustering, recurrent-convolutional models, and combinations of fuzzy time series. \citep{ref071,ref187,ref182} Research on temporal and graph Transformers, seasonal-trend models, and feature selection for prosumers further indicates the need to specify forecast horizons, covariates, and evaluation-data splits when comparing results. \citep{ref178,ref181,ref185,ref186,ref183,ref176} Some LLM forecasting methods also incorporate external factors to estimate generation or demand and explore zero-shot transfer and parameter-efficient adaptation. \citep{ref131,ref142,ref146,ref152} Their forecast quality likewise requires evaluation under the relevant task and data conditions.

For network structures, graph models retain electrical connections and line attributes. GNNs and related methods use these structures for state estimation, stability analysis, or risk assessment. \citep{ref021,ref158,ref160,ref168,ref134,ref162,ref167} State estimation, for example, infers operating quantities such as bus voltages from measurements. Measurement sufficiency is directly related to network connectivity, and a textual grid description alone can lose relations that determine observability and electrical feasibility. Other observations have their own representation requirements. Synchronized electrical waveforms require sampling, time synchronization, and channel-to-device mappings to be specified. Logs and alarms must be linked to the correct events, and inspection images must identify the observed object. Existing studies use vision foundation models for equipment inspection and image segmentation, \citep{ref155} photovoltaic graph models for spatiotemporal relations between sites, \citep{ref045} knowledge graphs for topology identification, \citep{ref088} and synchronized-waveform foundation models for electrical time series. \citep{ref006} These outputs can jointly explain the same phenomenon only when their objects, times, and operating conditions match.

Specialist models need to collaborate within the same research task. An LLM can coordinate their use according to research needs and organize their outputs into an explanation or report. Time-series models supply forecasts, graph models represent network connectivity, physical simulation or constrained optimization tests candidate feasibility, and knowledge graphs and retrieval systems preserve queryable evidence. Digital-twin modelling and energy-system reviews provide precedents for component integration, \citep{ref035,ref044,ref074,ref078,ref079,ref175} while multimodal knowledge-graph research in scientific discovery offers further references. \citep{ref205} A modular organization should retain each model's own validation rules, uncertainty statements, and refusal conditions. Even when a waveform, outage report, and simulation trajectory concern the same event, they have different noise sources, update frequencies, and evidentiary scope. If fusion retains only a single embedding and discards modality-specific checks, those differences become difficult to identify in subsequent interpretation.

This collaboration also requires connecting how the different materials are used. Research questions must be linked to retrieval results and approved assumptions to actual tool runs, while rejected assumptions and failed attempts are retained. Specialist-model interfaces and output provenance should be independently inspectable. The propagation of uncertainty into subsequent computations and expert approval of interpretations should be recorded with the relevant results. When modalities disagree, researchers can then trace the difference to data, assumptions, or models, preventing an LLM from combining incompatible evidence into an apparently consistent conclusion.

\section{Governance requirements and development roadmap}
Governance rules determine what an agent can do during research and how errors can be traced. Retrieving documents, executing code, and changing assumptions require different permissions, which must be distinguished in system design. Each information-processing operation or tool call should leave an inspectable record. Failures, approvals, and uncertainty statements should be linked to the corresponding results. Researchers can then determine which operations were authorized and which changes affected the final claim.

After diagnosis and validation, failure records can also support conclusions with explicit conditions. Non-convergence or constraint violations, for example, can inform analysis of the limitations of a model configuration, but cannot justify unverified feasibility or safety judgements. Incomplete assumptions or unclear applicability require corresponding limits on the conclusions. Explanations of failures and changes to critical assumptions require human review, and researchers should decide whether final conclusions are acceptable.

Governance must address LLM-system risks and grid operational-technology (OT) threats separately. Research on LLM integration discusses security risks and potential safeguards, \citep{ref003,ref004,ref149} while other studies examine attacks generated by non-LLM models such as generative adversarial networks (GANs) and autoencoders, \citep{ref147} or vulnerabilities in grid machine learning and state estimation themselves. \citep{ref111,ref159} LLMs, multimodal models, and other generative models have also been used for anomaly detection, \citep{ref068,ref136,ref117} but performance in detecting grid threats does not establish the security of a surrounding agent. The AI Scientist demonstrates a research system's ability to generate and execute code, \citep{ref052} making execution permissions a separate control requirement. This review recommends task-specific permissions and sandboxed code execution. Credentials should be protected and sensitive data isolated or anonymized. Review should examine prompt injection and data leakage, check the threat assumptions used in the power-system study, and establish that restricted operations can be revoked when necessary.

Some studies indicate that research quality also depends on whether explanations are faithful to the evidence and whether provenance is complete. \citep{ref135,ref040,ref056} Governance measures can accordingly be matched to operational risk. Formatting a report or extracting candidate variables can be automated while retaining logs. Determining model boundaries, authorizing sensitive-data access, interpreting protection failures, and deciding which conclusions can be published require review by the people responsible for those decisions. Approval points and audit records should identify the decision makers for subsequent examination.

Table~\ref{tab:roadmap} divides capability development into five stages. The earlier stages are already observable in existing applications: LLM assistants explain concepts and draft code, retrieval-based question answering attaches references or source links to answers, and tool-connected agents execute code and adjust operations from runtime feedback. Source-linked answers provide a starting point for traceability but still require citation checking and explicit assumptions. Grid-model generation and feedback-driven simulation demonstrate concrete applications of tool execution. \citep{ref014,ref120} The final two stages concern closed-loop experiments and collaboration across research tasks. Related systems have begun exploring these capabilities, \citep{ref052,ref103,seo2026spio} but further research is needed to establish how they can sustain experimental planning, handle failures, and support expert judgement in power systems. Capability expansion should proceed together with validation and permission management, preserving reviewable outputs and decision processes and identifying who approves and takes responsibility for critical operations.

\begin{table}[tbp]
\centering\small
\caption{Development roadmap for human-supervised AI researchers in power systems.}
\label{tab:roadmap}
\setlength{\tabcolsep}{4pt}\renewcommand{\arraystretch}{1.18}
\begin{tabularx}{\textwidth}{>{\hsize=0.7\hsize\linewidth=\hsize\raggedright\arraybackslash}X>{\hsize=1.1\hsize\linewidth=\hsize\raggedright\arraybackslash}X>{\hsize=1.3\hsize\linewidth=\hsize\raggedright\arraybackslash}X>{\hsize=0.9\hsize\linewidth=\hsize\raggedright\arraybackslash}X}
\toprule
\textbf{Stage} & \textbf{Main capability} & \textbf{Trustworthy research artifacts} & \textbf{Governance measures} \\
\midrule
LLM assistant & Explain concepts from literature and draft code or text for human use & Notes with sources and uncertainty statements, and draft code & Human checking before use \\
Traceable artifact builder & Extract source-linked evidence and represent assumptions as typed data structures & Source-supported claims and variable definitions with types and units & Citation, schema, and unit checks \\
Simulation-connected agent & Run generated cases and propose repairs from tool errors & Executable cases with corresponding solver, error, and validation records & Sandboxed execution and physical validators \\
Closed-loop research agent & Repeat simulations according to an experimental plan, compare results, and record negative cases & Experimental plans and complete run histories, including failures and comparison reports & Human approval of critical experimental decisions and research claims \\
Human-supervised AI researcher & Coordinate research tasks under expert authorization and use evidence and execution results in reporting & Research packages with sources, decision processes, and approval records & Expert decision authority, accountability, and institutional review \\
\bottomrule
\end{tabularx}
\end{table}

Research reporting can follow the evidence artifacts, modelling artifacts, execution records, and governance records in Table~\ref{tab:artifacts}, explaining the evidence, computation, and review supporting each conclusion. Reports should distinguish candidate claims proposed by an LLM from conclusions that have undergone physical, simulation, experimental, or expert review. They should provide the model and retrieval configurations, prompts, and tool versions needed for reproduction, together with scripts and logs, and describe failed attempts and human corrections. Where licensing or security conditions prevent full release of data or models, reports can provide data structures, conditions of applicability, and logs and execution descriptions available for controlled review. Research efficiency must be evaluated alongside the cost of checking. Given the electricity required for LLM training and inference, \citep{ref116} reports of task completion should also state the number of model calls, computational resource use, and human correction effort.

Whether LLMs improve power-system research must be established through comparative experiments on specific tasks. Tasks with sufficiently complete documents, data, and simulation interfaces can be used to compare results and workload with and without LLMs, recording computational resource use and human verification time and identifying the stages responsible for any efficiency gains. Experiments can deliberately introduce outdated clauses, unit errors, or missing measurements to observe whether the system detects problems and to trace how a retrieved item changes parameters, constraints, and final conclusions. When execution fails, evaluation should check whether repairs work, whether they change unapproved assumptions, and whether restricted tool requests can be blocked. Multimodal tasks should examine how conflicting evidence is handled and how uncertainty propagates into subsequent computations. Results can then identify which operations are suitable for automation and which changes require researchers to reconsider their judgement.

\section{Conclusions}
LLMs may reduce repetitive work in power-system research, including consulting literature, organizing assumptions, writing code, and coordinating simulations. This review examines power-system LLM applications, related scientific workflows, and physics-grounded power AI and proposes a knowledge-physics-simulation-governance framework. Evaluation should consider the retained research materials and actual validation process to determine whether a system helps researchers review conclusions, analyze failures, and revise judgements, alongside answer quality and task-completion speed.

Within this framework, LLMs translate information and coordinate tasks. Researchers retain authority over problem definition, data authorization, and critical assumptions and remain responsible for safety boundaries and final conclusions. Generated models should undergo task-appropriate equation and constraint checks and be tested by corresponding solvers or simulators. Sources, conditions of applicability, solver logs, negative results, and approval records should be retained in a research package so that efficiency gains rest on reviewable computations and judgements. The staged development roadmap and reporting requirements offer references for gradually expanding task scope and comparing different systems.

This review selects literature to develop a framework and does not provide an exhaustive systematic review or meta-analysis. Existing tasks and benchmarks are fragmented, and evidence from real engineering deployment remains limited. The proposed workflows and evaluation requirements therefore require testing through shared benchmarks, comparative experiments, and independent reproduction. Further research should identify the conditions under which different methods work, when expert intervention is necessary, and whether computational and human verification costs improve in practice.

\section*{Declaration of competing interest}
The authors declare that they have no known competing financial interests or personal relationships that could have appeared to influence the work reported in this paper.

\section*{Data availability}
No grid-operating data were generated in this study. This review is based on published literature and locally curated bibliographic records.

\FloatBarrier
\begingroup\small
\setlength{\bibsep}{4pt}

\endgroup
\end{document}